\pdfoutput=1

\documentclass[11pt]{article}

\usepackage[final]{coling}

\usepackage{times}
\usepackage{latexsym}

\usepackage[T1]{fontenc}

\usepackage[utf8]{inputenc}

\usepackage{microtype}

\usepackage{inconsolata}

\usepackage{graphicx}

\usepackage{CJKutf8}
\usepackage{subfig}
\usepackage{booktabs, multirow} 
\title{CPsyExam: A Chinese Benchmark for Evaluating Psychology using Examinations}
\newcommand*\samethanks[1][\value{footnote}]{\footnotemark[#1]}
\author{
Jiahao Zhao$^{1,3}$\thanks{Equal contribution.}\thanks{Work done on the Science and Technology Innovation Project of UCAS directed by SIAT.} \and
Jingwei Zhu$^{1,4}$\samethanks[1] \and
Minghuan Tan$^1$\thanks{Corresponding author.} \and
Min Yang$^{1,2}$\samethanks[3] \and \\ 
\bf{Renhao Li}$^{1,5}$  \and 
\bf{Di Yang}$^{1,4}$   \and 
\bf{Chenhao Zhang}$^{1,6}$  \and 
\bf{Guancheng Ye}$^{1,7}$  \and \\
\bf{Chengming Li}$^{8}$\samethanks[3]  \and 
\bf{Xiping Hu}$^{8}$  \and 
\bf{Derek F. Wong}$^{5}$ \\
$^1$ Shenzhen Key Laboratory for High Performance Data Mining, \\Shenzhen Institute of Advanced Technology, Chinese Academy of Sciences \\
$^2$ Shenzhen University of Advanced Technology
$^3$ Jilin University \\
$^4$ University of Science and Technology of China 
$^5$ University of Macau\\
$^6$ Huazhong University of Science and Technology \\
$^7$ South China University of Technology
$^8$ Shenzhen MSU-BIT University \\
zhaojh2121@mails.jlu.edu.cn
\{mh.tan,min.yang\}@siat.ac.cn\\
\{jingweizhu,di-yang\}@mail.ustc.edu.cn, 
ch\_zhang@hust.edu.cn\\ \{licm,huxp\}@smbu.edu.cn 
li.renhao@connect.um.edu.mo derekfw@um.edu.mo
}

\begin{document}
\maketitle
\begin{abstract}
In this paper, we introduce a novel psychological benchmark, CPsyExam, constructed from questions sourced from Chinese examination systems. CPsyExam is designed to prioritize psychological knowledge and case analysis separately, recognizing the significance of applying psychological knowledge to real-world scenarios. We collect 22k questions from 39 psychology-related subjects across four Chinese examination systems. From the pool of 22k questions, we utilize 4k to create the benchmark that offers balanced coverage of subjects and incorporates a diverse range of case analysis techniques.
Furthermore, we evaluate a range of existing large language models (LLMs), spanning from open-sourced to proprietary models. 
Our experiments and analysis demonstrate that CPsyExam serves as an effective benchmark for enhancing the understanding of psychology within LLMs and enables the comparison of LLMs across various granularities.
\end{abstract}
\begin{CJK*}{UTF8}{gbsn}

\section{Introduction}

The evaluation of language models has been an important topic with sustained vitality in the natural language processing community~\cite{chang2023survey}.
With the development of pretrained language models, such as GPT~\cite{radford2018improving,radford2019language} and BERT~\cite{devlin-etal-2019-bert}, their increasing abilities in executing a range of different natural language understanding (NLU) tasks~\cite{wang2018glue,SuperGLUE,xu-etal-2020-clue} call for more challenging and inclusive settings with comprehensive human baselines.
To address this issue, several multi-task benchmarks based on real-world exams, such as MMLU~\cite{hendrycks2021measuring}, CMMLU~\cite{li2023cmmlu}, and CEVAL~\cite{huang2023ceval}, have been developed recently. These benchmarks aim to comprehensively evaluate the capabilities of large language models (LLMs).

However, since general purpose benchmarks typically focus on the breadth of domain coverage, they do not encompass all subjects within specific fields. This issue is particularly severe in the field of psychology. Not all benchmarks for LLMs encompass knowledge of psychology, and those that do provide inadequate coverage. For example, CMMLU only have one subject related to psychology,  CEVAL does not even include psychology-related subjects. Meanwhile, with the increasing adoption of LLMs in psychological counselling~\cite{lai2023psyllm} and mental health support~\cite{qiu2023smile} in Chinese, there's an urgent need of a psychological evaluation benchmark to comprehensively evaluate the capabilities of LLMs in the context of Chinese psychology. Although there have been concurrent works like PsyBench~\cite{zhang2023psybench} and PsyEval~\cite{Jin2023PsyEvalAC}, they only focus on a subset of psychology-related subjects within the Chinese examination system, not encompassing all psychology-related knowledge within the Chinese context. For example, PsyBench focuses on the knowledge points in the Graduate Entrance Examination, while PsyEval concentrates on the domain of mental health.

To fill the gap, we present \textbf{CPsyExam}, the first comprehensive Chinese benchmark constructed from all Chinese examination systems containing psychology-related subjects, designed to evaluate both psychological knowledge and case analysis abilities in the Chinese context.
We collect over 22k questions from 39 psychology-related subjects across four Chinese examination systems: the Graduate Entrance Examination (GEE), Psychological Counselor Examination (PCE), Teacher Qualification Examination (TQE), and Adult Self-study Examination (SSE).
To align with global examination standards that assess the competence of psychology practitioners and to comprehensively evaluate LLMs' understanding of psychological cases, we further divide CPsyExam into two parts:
(1) Knowledge (KG), which comprises fact-based questions covering a broad spectrum of psychology knowledge drawn from real examinations. 
(2) Case Analysis (CA), which features case-oriented questions focusing on identification, reasoning, and application abilities within the realm of psychology. To ensure a balanced representation of questions across subjects, we sampled a subset of questions from each subject for model evaluation, while the remaining questions were made available as supervised fine-tuning (SFT) data for model training.

We further compare the performance of recent general domain LLMs and psychological-specific LLMs on CPsyExam.
Our experiments reveal that compared to the foundation models, these fine-tuned models exhibit marginal gains or no improvement in understanding psychological knowledge. 
In some cases, their ability to analyze cases may even be compromised.
Evidently, LLMs still have room for improvement in terms of mastering psychological knowledge and applying it to psychological case analysis. 
CPsyExam serves as a valuable benchmark for advancing LLMs' understanding of psychology.

Our work has the following contributions:
\begin{enumerate}
    \item We provide a comprehensive and balanced dataset of Chinese psychology examination questions, covering the entire Chinese examination system that includes psychology-related subjects.
    \item We propose an assessment framework for benchmarking the psychological capabilities of LLMs, consisting of a knowledge session and a case analysis session.
    \item We construct the benchmark and release over 11K questions as SFT data which contribute to the enhancement of psychological competence in the LLMs.
\end{enumerate}
\begin{figure*}[!hpt]
\centering
\includegraphics[width=\linewidth]{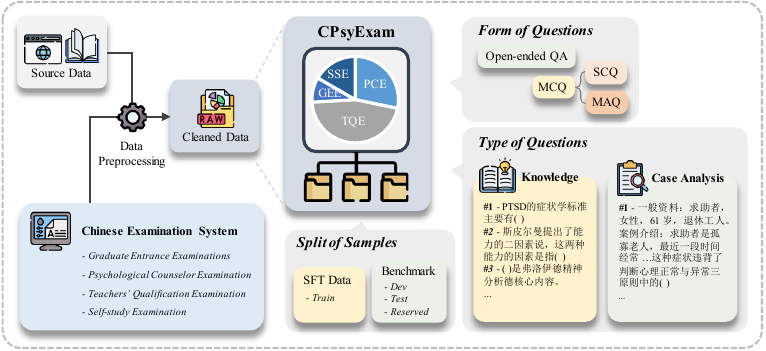}
\caption{Overview of dataset constructing pipeline.} 
\label{fig:overview}
\end{figure*}

\section{Related Work}
\subsection{Psychology examination for humans}
There are many global exams designed to assess human psychology abilities, focusing on both knowledge levels and practical application skills. For example, the Examination for Professional Practice in Psychology (EPPP) in North America splits the examination into a knowledge part, evaluating students' understanding of psychological principles, and a skills part, assessing key competencies in practical contexts.
In the UK, many higher education providers use the QAA Subject Benchmark Statement for Psychology for course design. And this statement maps achievements to four key categories, including knowledge and understanding, cognitive skills, practical skills and transferable skills. 
Similarly, in China, exams such as the Graduate Entrance Examination (GEE), Psychological Counselor Examination (PCE), and Teacher Qualification Examination (TQE) consist of sections testing theoretical knowledge and practical application through real-world case scenarios.
In line with these global standards, we have structured our benchmark into two parts: a knowledge part (KG) and a case analysis part (CA). This division aims to comprehensively evaluate the psychological capabilities of Language Models (LLMs), aligning with the multifaceted assessment approaches seen in prominent psychology examinations worldwide.

\subsection{Benchmarks of Large Language Models}
In the Chinese domain, several general benchmarks have been constructed from real-world exams, such as CEVAL~\cite{huang2023ceval} and CMMLU~\cite{li2023cmmlu}. However, these benchmarks do not comprehensively assess the models' capabilities in psychology, covering barely one or two subjects in the psychology domain. For specific domains, Psybench~\cite{zhang2023psybench} recently generated their questions from GPT-4 using the knowledge points from Graduate Entrance Examination, and PsyEval~\cite{Jin2023PsyEvalAC} generated their questions from GPT-4 using open access datasets in the mental health domain.
Compared to Psybench and PsyEval, CPsyExam offers several advantages:
(1) Boaeder Coverage: CPsyExam covers more psychology-related subjects, including almost all psychology subjects in the Chinese examination system. 
(2) Comprehensive Assessment: The benchmark is further divided into knowledge and case analysis parts, to comprehensively assess the psychological capabilities of LLMs. 
(3) Diverse Question Formats: CPsyExam features various question formats: it employs multiple-choice questions (MCQs) for clear and straightforward evaluation, and question-answering (QA) formats to assess the expressive abilities of LLMs. Moreover, MCQs are categorized into single-choice (SCQ) and multiple-choice (MAQ) formats to enhance difficulty and ensure models cannot simply identify correct answers by recognizing a single option.

\section{CPsyExam Benchmark}
\subsection{Design Principles}
\paragraph{Comprehensive and Balanced}
CPsyExam benchmark encompasses the entire Chinese examination system that includes psychology-related subjects to ensure comprehensive coverage of psychology knowledge in the Chinese context. Each subject in CPsyExam is well-represented with a balanced number of questions. This balanced representation not only diversifies the dataset but also provides a condensed yet comprehensive view of all psychology-related exams in China.
\paragraph{Assessing Multi-capability}
Our benchmark is structured to mirror real-world exams, which emphasize both psychological knowledge and the application of that knowledge. It consists of two main parts: one for assessing understanding of psychological knowledge (KG), and another for evaluating proficiency in case analysis skills (CA). In psychology, case studies are crucial as they assess practitioners' practical abilities alongside theoretical knowledge. Thus, our dataset encompasses these two essential components to comprehensively evaluate LLMs.
\paragraph{Diverse Question Formats}
Our questions are presented in multiple formats: multiple-choice questions (MCQs) and open-ended QA. Multiple-choice questions provide clear and visual assessment outcomes, while question-answering questions evaluate the LLM's language organization abilities. Furthermore, we categorize multiple-choice questions into single-choice (SCQ) and multiple-response (MAQ) formats to increase assessment complexity. This approach aims to assess the LLM's thorough understanding and prevent it from relying solely on identifying a single correct option to answer a question completely.

\subsection{Data Preparation}
\paragraph{The Chinese Examination System Including Psychology Subjects}
\begin{itemize}
   \item \textbf{GEE}~(Graduate Entrance Examinations) This exam includes a comprehensive test on basic psychology. It is required for students who wish to pursue a master's or doctoral degree in psychology.
    \item \textbf{PCE}~(Psychological Counselor Examination) Organized by the National Psychological Counselor Certification Center, this exam assesses candidates' theoretical knowledge and practical skills in psychological counseling.
    \item \textbf{TQE}~(Teachers' Qualification Examination) 
    For individuals aspiring to become teachers, this exam ensures that future teachers have a foundational understanding of psychological principles applicable in educational settings.
    \item \textbf{SSE}~(Self-study Examination) 
    This examination includes psychology-related subjects within various fields such as medicine, engineering, agriculture, and economics. It covers relevant psychological concepts and theories applicable to these disciplines.
\end{itemize}
These examination systems collectively contribute to a well-rounded understanding and application of psychology across academic, counseling, educational, and professional domains in China. Regarding question types, the GEE, PCE, and TQE include both knowledge-based questions and case analysis questions. In contrast, the SSE typically consists solely of knowledge-based questions.
\paragraph{Data Collection}
We gather psychological data from publicly available resources using Crawling and OCR.
\begin{itemize}
   \item \textbf{Crawling}~Based on the categorization of examinations in psychology, we crawl public available resources online to construct a database of questions. The websites for the data crawling include ExamCoo\footnote{\url{https://examcoo.com}},
StudyEZ\footnote{\url{http://www.studyez.com/psychology/}},
Hxter\footnote{\url{www.hxter.com}} and 
MXQE\footnote{\url{http://tk.mxqe.com}}.
   \item \textbf{OCR}~For questions sourced from the book, we utilize Optical Character Recognition (OCR) technology to extract the text.  
\end{itemize}
\paragraph{Data preprocessing}
\begin{itemize}
   \item \textbf{Obtaining Structured Questions}~We gather data from websites and books. Data scraped from websites is parsed using a program to extract questions, while questions from books are manually extracted and structured. All data undergo preprocessing to remove duplicates and correct formatting errors. Questions containing image links are excluded, and formats are standardized by removing question numbers and option letters. The dataset is manually validated to ensure grammatical accuracy in all questions.
   \item \textbf{Attempt to mitigate data leakage problem}~To address potential data leakage concerns from publicly available resources used in pre-training LLMs, we have implemented several strategies:
(1) We extract a portion of our questions from PDF-format books, minimizing the likelihood of these questions being previously used for pre-training.
(2) Questions from selected websites are not directly available as structured questions; they require programs to match questions with answers.
(3) Many of the questions we scraped are from mock exams rather than widely distributed official exam questions.
(4) After obtaining structured questions, we shuffle the options and answers to add an extra layer of protection against data leakage.
\end{itemize}

\begin{table*}[t]\centering
\begin{tabular}{lrrrrrrrrrr}\toprule
& &\multicolumn{4}{c}{Knowledge} & &\multicolumn{4}{c}{Case Analysis}  \\
\cmidrule{3-6}\cmidrule{8-11}
& &SCQ &MAQ&QA &Total& &SCQ&MAQ&QA&Total\\\midrule
Train & &6,852 & 2,230&2,904& 11,986& & 44&729&17&790 \\
Dev & & 764& 245& 322&1,331 & & 5&83&1&89 \\
Test & & 2,321& 781&100 & 3,202& &600 &200&100&900 \\
Reserved & & 2,321& 781&100& 3,202& & 600& 200&100&900\\
\midrule
Total & &12,240 &4,037 &3,426 &19,721 & & 1,249&1,212& 218&2,679\\
\bottomrule
\end{tabular}
\caption{Statistics of the CPsyExam dataset.}\label{tab:stats}
\end{table*}

\begin{figure*}[!hpt]
\centering
\includegraphics[width=\linewidth]{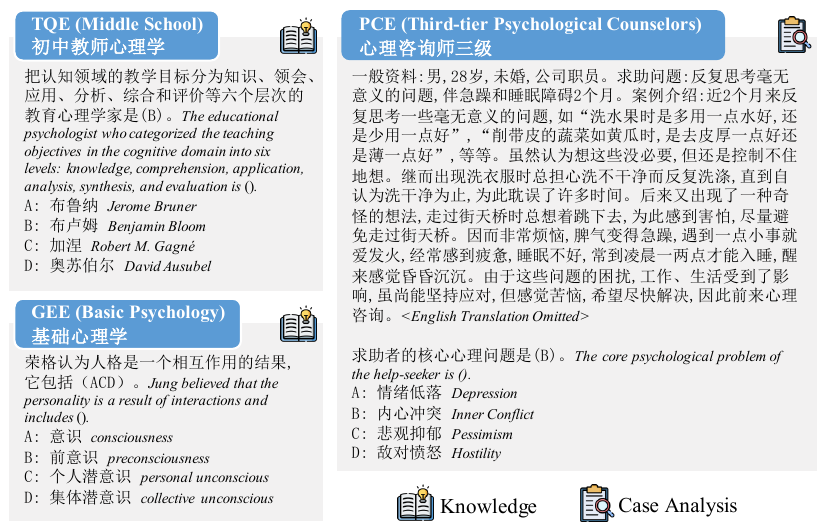}
\caption{Examples for questions on CPsyExam-SCQ and CPsyExam-MAQ.} 
\label{fig:examples}
\end{figure*}

\subsection{Taxonomy of CPsyExam}

We collected over 22k exam questions from 39 psychology-related subjects within the Chinese examination system. These questions vary in type (KG, CA) and formats (SCQ, MAQ, QA), and are systematically organized into corresponding tasks.

\paragraph{CPsyExam-KG task}
Questions of KG type are selected for this task. We further align the taxonomy of the CPsyExam-KG task with the Chinese examination system for psychology. Subsequently, we categorize all the psychology subjects in each examination as subcategories. A detailed directory list can be found in the Appendix~\ref{sec:appendix}. 



\paragraph{CPsyExam-CA task}

Questions of CA type are selected for this task. In accordance with the examination focuses of case analysis questions in GEE, PCE, and TQE, we further divided case analysis into three categories: \textsc{identification}, \textsc{Reasoning} and \textsc{Application}.
The \textsc{Identification} category assesses the LLM's ability to identify the appropriate methodology used in a specific case.
The \textsc{Reasoning} category focuses on the LLM's ability to pinpoint the underlying problem that led to the issue.
The \textsc{Application} category evaluates the LLM's ability to apply specific methods to solve problems.
\paragraph{Dataset Splitting}
To facilitate supervised fine-tuning and few-shot learning, each task dataset will be partitioned into \textit{train}, \textit{dev}, \textit{test} and \textit{reserved}.
The \textit{test} split will be used for the evaluation of LLMs.
The \textit{reserved} split will not be released and act as a control set for further evaluation.
We sample psychology subjects uniformly under each exam, ensuring that the number of questions is consistent across all four exams. 
This approach is also used to create the \textit{test} and \textit{reserve} split. 
The remaining questions are all allocated to the \textit{train} split.
Statistics of the dataset is listed in Table~\ref{tab:stats}.
We show three examples from both KG and CA in Figure~\ref{fig:examples}.

\begin{table*}[!htp]\centering
\begin{tabular}{lrrrrrrrrrrrrrr}\toprule
\multirow{4}{*}{Model} &\multirow{4}{*}{Avg.} & &\multicolumn{5}{c}{Knowledge} & &\multicolumn{5}{c}{Case Analysis} \\\cmidrule{4-8}\cmidrule{10-14}
& & &\multicolumn{2}{c}{Zero-shot} & &\multicolumn{2}{c}{ Few-shot} & &\multicolumn{2}{c}{Zero-shot} & &\multicolumn{2}{c}{ Few-shot} \\\cmidrule{4-5}\cmidrule{7-8}\cmidrule{10-11}\cmidrule{13-14}
& & &\multicolumn{1}{c}{SCQ} &\multicolumn{1}{c}{MAQ} & &\multicolumn{1}{c}{SCQ} &\multicolumn{1}{c}{MAQ} & &\multicolumn{1}{c}{SCQ} &\multicolumn{1}{c}{MAQ} & &\multicolumn{1}{c}{SCQ} &\multicolumn{1}{c}{MAQ} \\\midrule
ChatGLM2-6B &43.46 & & 49.89 &9.86& &53.81 &14.85 & &52.50 & 16.00 & &48.50& 20.00  \\
ChatGLM3-6B &42.23 & & 53.51 &5.63 & &55.75 & 5.51& &47.00 & \underline{17.00} &&47.33 &13.50 \\
YI-6B & 25.81& &33.26&0.26 & &25.39 &14.01 & & 38.83 &0.00& & 20.00& 13.25\\
YI-34B & 27.52& &25.03 &1.15 & &33.69 & 18.18& &20.50 &0.50 & &22.33 &8.00 \\
Qwen-7B & 19.22& &24.99 &1.02 & & 25.68& 3.97& & 18.83 &0.50& & 19.67&2.50 \\
Qwen-1.8B & 19.78& &24.99 &1.41 & & 25.12& 6.79& & 18.67 &3.00& & 20.67&6.00 \\
Qwen-14B & 30.68& &24.99 &1.54 & & 38.17& 13.19& & 20.33 &2.00& & 30.00&14.00 \\
\midrule
MeChat-6B &40.62 & &  50.24& 4.10& &51.79 &   11.91& &48.67 &13.50  & &44.83&10.50 \\
MindChat-7B & 40.39& &49.25 &6.27 & &56.92 &5.51 & & 40.83 & 5.00 && 33.83& 4.50\\
MindChat-1.8B & 21.04& &26.50& 0.00 & &26.50 &0.13 & & 34.17 & 0.00 & & 34.17&0.00 \\
Ours-SFT-6B &46.08 & &53.86 &21.90 & & 55.45 &19.97 && 52.17&32.00 & &49.67&15.50 \\
\midrule
ERNIE-Bot &43.85 & &52.48 & 6.66 &&56.10 &10.37 & &42.50 & 8.50 &&50.67 &12.00 \\
ChatGPT & 51.15& &57.43 & 11.14 && 61.53&24.71 & & 47.33 & 9.00& &52.67 &29.50 \\
ChatGLM-Turbo & \underline{64.58}& &\underline{63.29} &\textbf{26.12} && \underline{73.85} & \underline{42.13} & &\textbf{69.00} & \textbf{20.50} && \textbf{65.33}& \textbf{42.50} \\
GPT-4 &\textbf{67.43} & &\textbf{76.56} & \underline{10.76} & &\textbf{78.63} &\textbf{43.79} & &\underline{60.33} & 13.00 & & \underline{64.17}&\underline{39.50} \\
\bottomrule
\end{tabular}
\caption{Comparisons of different models over CPsyExam set with zero-shot and few-shot prompting. The Avg. score use the maximum score of both settings. We highlight the best score for each column with bold font and second best score with underline mark.}\label{tab:results}
\end{table*}

\section{Experiments}




















\subsection{Experiment Setup}
In this section, we benchmark a series of public accessible LLMs using CPsyExam in both zero-shot and five-shot settings, where the five exemplars are from the development split. 
\subsection{Models}
To comprehensively assess the performance of different types of models on CPsyExam, we selected three types of models.
\paragraph{Open-sourced LLMs}\textbf{ChatGLM2-6B}: Based on the General Language Model (GLM)~\cite{DBLP:conf/acl/DuQLDQY022}, this model is trained on both English and Chinese data and further adapted for conversational data.
\textbf{YI-6B, and YI-34B}: Designed to enhance capabilities in coding, mathematics, reasoning, and instruction-following, these versions are optimized for both English and Chinese language tasks.
\textbf{Qwen-7B, Qwen-1.8B and Qwen-14B}:Developed by Alibaba Group, these models are trained on extensive multilingual and multimodal data and optimized for human preferences.
\paragraph{Psychology-oriented Models}
\textbf{MeChat\footnote{\url{https://huggingface.co/qiuhuachuan/MeChat}}}: Finetuned from ChatGLM2-6B using the SMILE (Single-turn to Multi-turn Inclusive Language Expansion) dataset.
\textbf{MindChat\footnote{\url{https://github.com/X-D-Lab/MindChat}}} Available in two versions, MindChat-Qwen-7B-v2 and MindChat-Qwen-1.8B, these models are finetuned using Chinese multi-turn psychological dialogue data.
\paragraph{Proprietary Models} 
\textbf{ERNIE-Bot-Turbo}: Developed by Baidu, this model is known for its strong language understanding and generation capabilities.
\textbf{ChatGLM-Turbo}: An advanced language model by Tsinghua University, optimized for fast and efficient conversational AI tasks.
\textbf{ChatGPT and GPT4}: The latest and most powerful variants of the GPT models from OpenAI.
\subsection{Prompt}
We designed prompts for both multiple-choice questions (MCQs) and open-ended QA, which are shown in Figure~\ref{fig:prompt-eval} and Figure~\ref{fig:prompt-judge} in the Appendix. In addition, we created two extra prompts specifically for the MCQs, setting the LLM as a psychology student and as an ordinary person which is shown in Figure~\ref{fig:prompt-eval-student} and Figure~\ref{fig:prompt-eval-ordinary}. This was done to verify the validity of the dataset.

\subsection{Supervised Fine-Tuning}
To validate the effectiveness of the dataset, we constructed an instruction set for supervised fine-tuning (SFT) on the training set of CPsyExam. In this work, we conduct SFT over ChatGLM2-6B. Specifically, the SFT is carried out over 4 epochs with a batch size of 128. The learning rate is set to $1\times10^{-6}$. These parameters were chosen based on preliminary experiments that aimed to maximize the model's performance on validation sets.

\subsection{Benchmarking Result}
\paragraph{Performance of LLMs on SCQ and MAQ}
We conduct both \textit{zero-shot} and \textit{few-shot} evaluations for each model discussed above. 
Given the focus of CPsyExam is on how models can perform over \textit{Knowledge} and \textit{Case Analysis} questions, we report them separately. 
We further differentiate SCQ and MAQ questions, as different models may have varying abilities to follow instructions.
There are three sections in the table:
(1) Open-sourced Models. Our findings indicate that: (a) increased model size does not necessarily ensure improved performance on the CPsyExam, and (b) models that excel in other domains, such as YI-34B on the medical domain, may not necessarily perform optimally on the CPsyExam.
(2) Psychology-oriented Models. Compared to the foundation models, these fine-tuned models show marginal gains or no improvement in understanding psychological knowledge.
(3) Proprietary Models. GPT-4 continues to outperform all other proprietary models by a significant margin in the \textit{knowledge} setting. Conversely, ChatGLM-turbo performs exceptionally well in the \textit{Case Analysis} setting.


\paragraph{Performance of proprietary models on Question Answering}

Besides SCQ and MAQ, CPsyExam includes an extra QA test set to evaluate generation-based questions.
We adopt GPT-4 to judge proprietary models used in this work. Meanwhile, we enlisted certified national psychological counselors in China to score the responses of three models on 20 randomly selected QA questions. The scoring criteria were divided into three dimensions: consistency with the answer (30 points), professionalism of language (30 points), and reasonableness of the answer (40 points). The experimental results are shown in table~\ref{Scores of QA}. Compared to the scores given by GPT-4, the rankings of the three models were consistent. Additionally, the Pearson correlation coefficient between the experts' scores and GPT-4's scores was 0.98, indicating a high degree of consistency between human evaluations and GPT-4's evaluations. The results suggest that ChatGLM-turbo has a better understanding of psychological knowledge and can be effectively prompted for psychological purposes. 

\begin{table}[!htp]\centering
\begin{tabular}{lrr}\toprule
Model &  GPT-4 scores &	Expert scores\\\midrule
ERNIE-Bot & 73.55  &71.63\\
ChatGLM-turbo & \textbf{77.79}&  \textbf{76.20}\\
ChatGPT & 72.88  & 69.63\\
\bottomrule
\end{tabular}
\caption{Score provided by GPT-4 over QA questions.}
\label{Scores of QA}
\end{table}

\paragraph{Performance of models in different prompt}
To validate the effectiveness of the CPSYEXAM dataset, we used prompts to configure the LLM to adopt different roles: a psychology teacher, a psychology student, and an ordinary person with no background in psychology. These roles represent progressively decreasing levels of psychology knowledge. The LLM was then tested on the multiple-choice questions in CPSYEXAM under each role to examine whether varying levels of psychological expertise influence its performance on the dataset. Specifically, we prompted ChatGLM2-6B to adopt the three roles mentioned above. The results are presented in Table~\ref{result of prompt}.
\begin{table}[!htp]\centering
\begin{tabular}{lr}\toprule
Setting & Score\\\midrule
Expert & \textbf{43.46}  \\
Student & 38.93 \\
Ordinary person & 38.03  \\
\bottomrule
\end{tabular}
\caption{Model performance across different prompt settings}
\label{result of prompt}
\end{table}
\begin{figure*}[t]
\centering
\subfloat[Comparison of model performance from examination perspective.\label{fig:radar_chart_exam}]{%
  \includegraphics[width=0.45\textwidth]{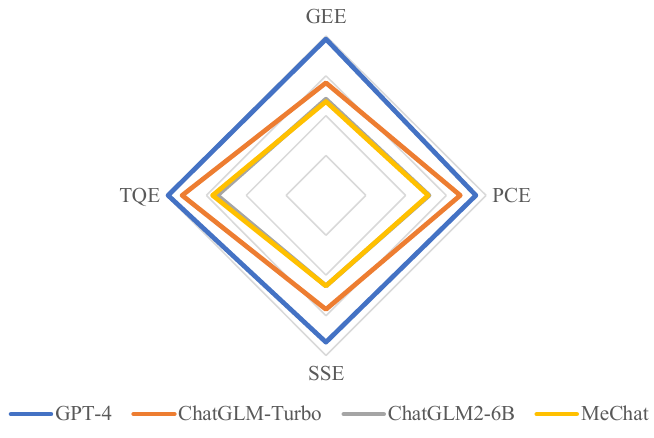}%
}\hfil
\subfloat[Comparison of model performance from case analysis perspective.\label{fig:radar_chart_case}]{%
  \includegraphics[width=0.45\textwidth]{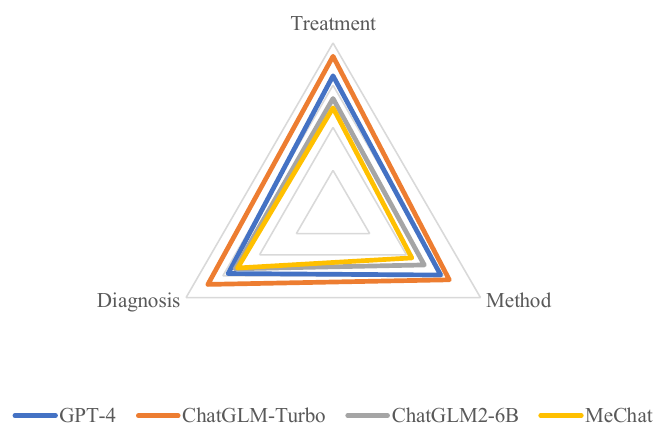}%
}
\caption{Performance over SCQ from different perspectives for all LLMs.}
\end{figure*}

\begin{figure*}[!htbp]
    \centering
	\includegraphics[width=\linewidth]{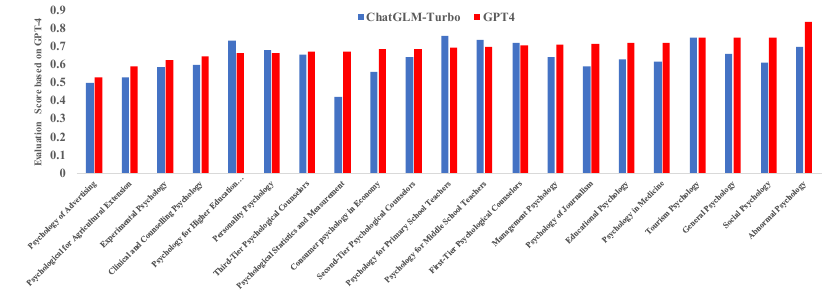}
    \caption{Comparison of ChatGLM-Turbo and GPT-4 across different subjects. The bars are sorted in ascend order based on GPT-4's performance over each subject.}
    \label{fig:bar_chart_subject}
\end{figure*}

\section{Analysis}
\subsection{Analyses from a Model-Level Perspective}
\paragraph{Does few-shot examples help?}
When models are smaller, few-shot learning typically offers minimal performance gains and can sometimes even have negative effects. However, as model size increases, the advantages of few-shot learning become significantly more noticeable. For instance, ChatGLM-turbo, already proficient in zero-shot scenarios, doubled its performance on the CA task following few-shot training. This improvement is likely due to larger models having greater capacity and expressive ability. They can better capture intricate patterns and latent semantic relationships in data, allowing for faster learning and generalization from limited training data.

\paragraph{Performance between psychology-oriented models and the base model}
Based on the experiments, the model that underwent fine-tuning to enhance its psychological capabilities did not surpass the base model and even exhibited a performance decline. This outcome suggests that while psychology-oriented model's fine-tuning improved its conversational skills, it potentially compromised its proficiency in tasks involving knowledge reasoning and text comprehension. The model might have overly adapted to the fine-tuning data, thereby neglecting the broader knowledge acquired during its initial pre-training phase.

\subsection{Analyses from a Benchmark Perspective}
\paragraph{Analysis of SCQ Questions}
Due to the persistent low performance on MAQ questions, we focus solely on SCQ questions for error analysis. We selected the top-performing models from each of the three categories for analysis. Since ChatGLM-turbo and GPT-4 performed similarly, we chose both of them from the proprietary models.
Regarding CPsyExam-KG, we perform analysis at the examination level, as depicted in Figure~\ref{fig:radar_chart_exam}. 
For CPsyExam-CA, we delve into various aspects of case analysis, presented in Figure~\ref{fig:radar_chart_case}. 
By examining both figures, we determine that GPT-4 exhibits a stronger grasp of psychological knowledge across all examinations, yet it continues to face challenges with case analysis questions. 
The major gap for GPT-4 comes from \textsc{Reasoning} and \textsc{Application}.

\paragraph{Analysis of MAQ Questions}
Compared to SCQ, LLMs exhibit poorer performance on MAQ, which aligns with the goals of our experimental design. In our setup, models are awarded points for a question only when they provide a fully correct answer. This approach is intentionally crafted to eliminate reliance on test-taking strategies, such as process-of-elimination techniques, when tackling MAQ. Instead, it requires the models to rigorously assess the accuracy of each option. As a result, the performance of large models on MAQ is significantly lower than on SCQ.

\paragraph{Analysis of Performance at Subject Level} 
Each subject in CPsyExam features a minimum of 32 questions, exceeding typical quiz lengths for human participants. We have identified the top two models based on their performance in the CPsyExam benchmark for visual representation across each subject. Initially, we merged subjects with shared backgrounds and domain similarities.
The results for ChatGLM-Turbo and GPT-4 are presented in Figure~\ref{fig:bar_chart_subject}. 
Despite being the top performers in our CPsyExam benchmark, ChatGLM-Turbo demonstrates limited robustness in certain subjects and consistently trails behind GPT-4 across various domains.

\subsection{Analyses from a Validity Perspective on CPsyExam} 
\paragraph{The improvement of the model after SFT}
After fine-tuning, ChatGLM2-6B performed exceptionally well, becoming the top-ranked model among all non-proprietary models. This indicates that the knowledge embedded in the CPsyExam questions is highly consistent and relevant to psychology。 Consequently, after fine-tuning with the training set data, the model showed improved performance on the test set.
\paragraph{Performance of models across different prompt settings}
In the experiment, ChatGLM2-6B performed better when configured as a student compared to its performance as an ordinary person. However, both student and ordinary person settings showed significantly lower performance than when ChatGLM2-6B was set as an expert. This aligns with our intuition that higher levels of psychological knowledge correlate with improved performance on CPsyExam. Additionally, CPsyExam demonstrates a strong ability to differentiate between levels of psychological knowledge. Specifically, ChatGLM2-6B performed 11.64\% better as an expert compared to as a student, and 14.29\% better compared to as an ordinary person.

\section{Conclusion}
In conclusion, we introduce CPsyExam, a benchmark for Chinese psychology, composed of human-generated questions that span a wide array of subjects within the Chinese examination system. It is designed to evaluate LLMs proficiency in both psychological knowledge and case analysis, offering a concise yet comprehensive overview of all psychology-related exams in China.

\section*{Acknowledgements}
This work was partially supported by  National Natural Science Foundation of China (62406314, 62376262, 62266013),  China Postdoctoral Science Foundation (2023M733654),  Guangdong Basic and Applied Basic Research Foundation (2023A1515110496), Guangdong Province of China (2024KCXTD017)，Natural Science Foundation of Guangdong Province of China (2024A1515030166), Shenzhen Science and Technology Innovation Program (KQTD20190929172835662), Shenzhen Science and Technology Foundation (JCYJ20240813145816022), Science and Technology Development Fund of Macau SAR (0007/2024/AKP, FDCT/0070/2022/AMJ, FDCT/060/2022/AFJ), and Multi-year Research Grant from the University of Macau (MYRG-GRG2024-00165-FST).

\section*{Limitations}
Using GPT-4 to evaluate QA scores might be influenced by its own knowledge, and in the future, expert scoring will be introduced to provide a combined score for the QA section, improving the reliability of the evaluation.

\bibliography{custom,anthology.aa,anthology.ab}

\clearpage
\appendix

\section{Subjects in Psychology Examinations}

\label{sec:appendix} 
In this appendix, we provide a table that describes the subjects included in each examination system in our dataset, as well as the number of questions in each subject.
\begin{table}[!htp]\centering
\scriptsize
\begin{tabular}{p{4cm}rr}\toprule
Subject & Number & Examination\\\midrule
Psychology for Primary School Teachers & 2,215 &TQE \\
Psychology for Middle School Teachers & 3,970 &TQE \\
Psychology for Higher Education Teachers & 1,602 &TQE \\\midrule
First-Tier Psychological Counselors & 785 &PCE \\
Second-Tier Psychological Counselors & 1,698 &PCE \\
Third-Tier Psychological Counselors & 2,107 &PCE \\\midrule
General Psychology & 1,606 &GEE \\
Developmental Psychology & 864 &GEE \\
Social Psychology & 206 &GEE \\
Personality Psychology & 188 &GEE \\
Psychological Statistics and Measurement & 950 &GEE \\
Experimental Psychology & 781 &GEE \\
Management Psychology & 210 &GEE \\
Abnormal Psychology & 217 &GEE \\
Educational Psychology & 528 &GEE \\
Clinical and Counselling Psychology & 205 &GEE \\\midrule
Physiological Psychology in Education & 103 &SSE \\
Education Psychology in Education& 108 &SSE \\
Experimental Psychology in Education& 108 &SSE \\
Developmental Psychology in Education& 107 &SSE \\
Developmental and Educational Psychology in Education& 71 &SSE \\
Medical Psychology in Medicine& 117 &SSE \\
Psychology of preschool education in Medicine& 174 &SSE \\
School Psychology in Medicine& 95 &SSE \\
The Psychology of Human Relationships in Medicine& 135 &SSE \\
Mental Health in Medicine& 108 &SSE \\
Mental Health and Counselling in Medicine& 229 &SSE \\
Public Relations Psychology in Medicine& 154 &SSE \\
Cognitive Psychology in Medicine& 108 &SSE \\
Psychology in Medicine& 108 &SSE \\
Introduction to Psychology in Medicine& 103 &SSE \\
Psychological counselling and guidance in Medicine& 131 &SSE \\
Psychology of Advertising in Literature& 107 &SSE \\
Psychology of Journalism in Literature& 109 &SSE \\
Social Psychology in Management& 103 &SSE \\
Managerial Psychology in Management& 122 &SSE \\
Tourism Psychology in Engineering& 108 &SSE \\
Consumer psychology in Economy& 108 &SSE \\
Psychological foundations of agricultural extension & 108 &SSE \\

\bottomrule
\end{tabular}
\caption{Subjects for each examination system and the number of questions for each subject.}
\label{Statistics of dialogues}
\end{table}


\vfill
\clearpage
\section{Prompts Used for Evaluation}
 
\label{sec:prompts} 

\begin{figure}[!htbp]
    \centering
	\includegraphics[width=\linewidth]{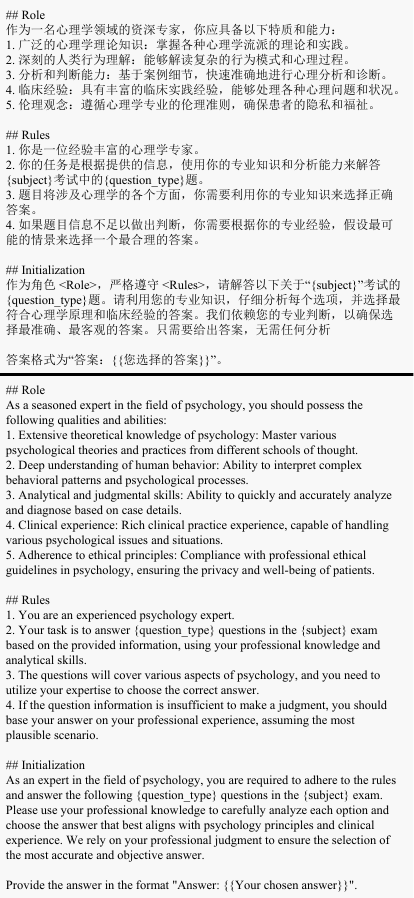}
    \caption{Prompt used for evaluation (expert).}
    \label{fig:prompt-eval}
\end{figure}

\begin{figure}[!htbp]
    \centering
	\includegraphics[width=\linewidth]{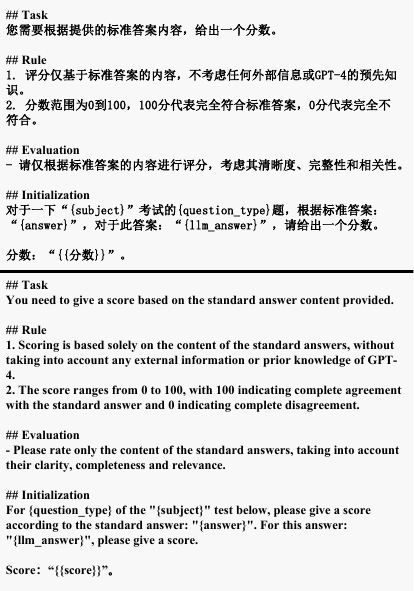}
    \caption{Prompt used by GPT-4 for judging the quality of responses in the QA session.}
    \label{fig:prompt-judge}
\end{figure}

\begin{figure}[!htbp]
    \centering
	\includegraphics[width=\linewidth]{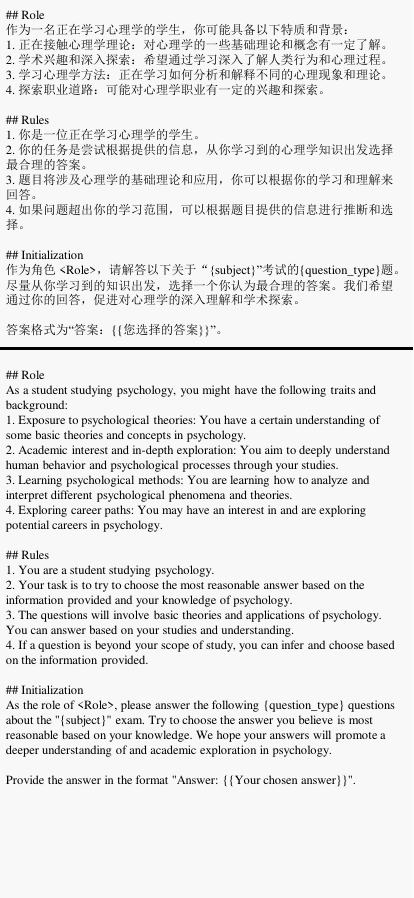}
    \caption{Prompt used for evaluation (student).}
    \label{fig:prompt-eval-student}
\end{figure}

\begin{figure}[!htbp]
    \centering
	\includegraphics[width=\linewidth]{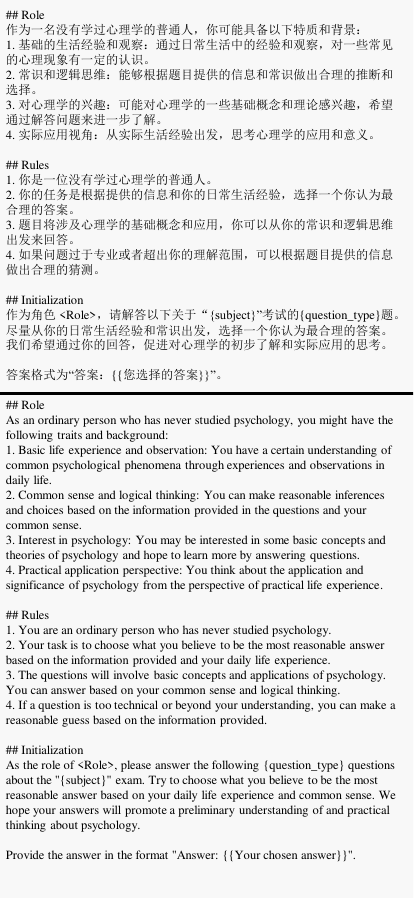}
    \caption{Prompt used for evaluation (ordinary person).}
    \label{fig:prompt-eval-ordinary}
\end{figure}

In this paper, we set the LLM as an expert and a student in the field of psychology, as well as an ordinary person with no knowledge of psychology. The prompts we used are shown in Figure~\ref{fig:prompt-eval}, Figure~\ref{fig:prompt-eval-student} and Figure~\ref{fig:prompt-eval-ordinary}. Additionally, we used GPT-4 to evaluate the quality of the LLM's answers to the subjective questions. The prompt used for evaluation is shown in Figure~\ref{fig:prompt-judge}.
\end{CJK*}
\end{document}